\documentclass[graybox]{svmult}

\usepackage{type1cm}        
\usepackage{makeidx}         
\usepackage{graphicx}        
\usepackage{multicol}        
\usepackage[bottom]{footmisc}

\usepackage{newtxtext}       %
\usepackage{newtxmath}       
\usepackage[numbers]{natbib}
\usepackage{float}           
\usepackage{amsmath}
\usepackage{float}
\usepackage{siunitx}

\usepackage{xcolor}     
\usepackage{ntheorem}   
\usepackage{mdframed}   

\theoremstyle{nonumberplain}
\theoremheaderfont{\bfseries}
\newmdtheoremenv[
linecolor=gray,leftmargin=60,
rightmargin=40,
backgroundcolor=gray!40,
innertopmargin=0pt,
ntheorem]{myprop}{CNNs:}[section]

\makeindex             

\begin{document}

\title*{Generative Adversarial Networks}
\author{Gilad Cohen and Raja Giryes}
\institute{Gilad Cohen \at Tel Aviv University, Tel Aviv 6997801, \email{giladco1@post.tau.ac.il}
\and Raja Giryes \at Tel Aviv University, Tel Aviv 6997801 \email{raja@tauex.tau.ac.il}}
%
%
\maketitle

\abstract{Generative Adversarial Networks (GANs) are very popular frameworks for generating high-quality data, and are immensely used in both the academia and industry in many domains. Arguably, their most substantial impact has been in the area of computer vision, where they achieve state-of-the-art image generation. This chapter gives an introduction to GANs, by discussing their principle mechanism and presenting some of their inherent problems during training and evaluation. We focus on these three issues: (1) mode collapse, (2) vanishing gradients, and (3) generation of low-quality images. We then list some architecture-variant and loss-variant GANs that remedy the above challenges. Lastly, we present two utilization examples of GANs for real-world applications: Data augmentation and face images generation.}

\section{Introduction to GANs}
\label{sec:introduction_to_gans}
Generative adversarial networks (GANs) are currently the leading method to learn a distribution of a given dataset and generate new examples from it. 
To begin to understand the concept of GANs, let us consider an interplay between the police and money counterfeiters. A state just launched its new currency and millions of bills and coins are widespread all over the country. Recently, the police detected a flood of counterfeit money in circulation. Further inspection reveals that the forged bills are lighter than the genuine bills, and can be easily filtered out. The criminals then find out about the police's discovery and use new printers which control better the cash weight. In turn, the police investigate and discover that the new counterfeit bills have a different texture near the corners, and remove them from the system. After many iterations of generating and discriminating forged money, the counterfeit money becomes almost indistinguishable from the real money. Note that the system has two agents: 1) The counterfeiters who create "close to real" money and 2) The police who detect counterfeit bills adequately.

Back to deep learning: The above two agents are called "generator" and "discriminator", respectively. These two entities are trained jointly, where the generator learns how to fool the discriminator with new adversarial examples out of the dataset distribution; and the discriminator learns to distinguish between real and fake data samples.
The GAN architecture is used in more and more applications since its introduction in 2014. It was proved successful in many domains such as computer vision \cite{Dziugaite2015TrainingGN,karras2018progressive,hussein2019imageadaptive,SRGAN}, semantic segmentation \cite{Luc2016SemanticSU,Pix2Pix,nvidia_synthetic_GAN,Hoffman2019CyCADA}, time-series synthesis \cite{Brophy2019QuickAE,Hartmann2018EEGGANGA}, image editing \cite{rottshaham2019singan,lample2017fader,gal2021stylegannada,Abdal21StyleFlow,xia2021gan}, natural language processing \cite{MaskGAN,JetchevBV16,LeakGAN}, text-to-image generation \cite{ramesh2021zeroshot,radford2021learning,patashnik2021styleclip}, and many more. In the next section, we depict the basic GAN architecture and loss. Later, we will present more sophisticated architectures, losses, and common usages.

\section{The basic GAN concept}
\label{sec:concept_of_basic_gan}

\begin{figure}[h]
\centering
\includegraphics[width=\linewidth]{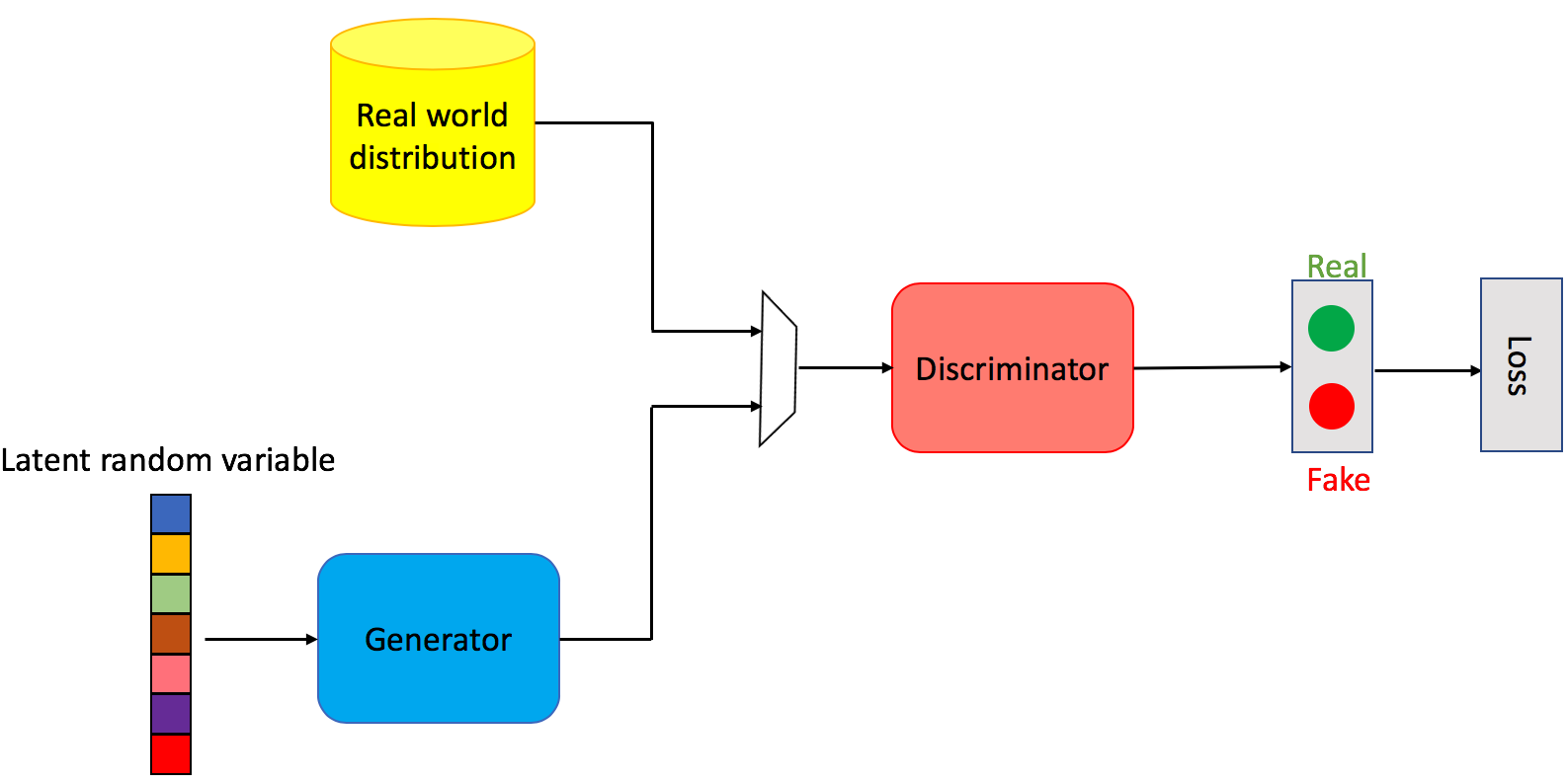}
\caption{Basic GAN structure. The discriminator and generator are two deep neural networks trained jointly. The discriminator is trained for the task of classifying whether an input image is natural (real) or generated (fake), while the generator is optimized to fool the discriminator.}
\label{fig:GAN_basic}
\end{figure}

The first GAN that was introduced by \citet{GAN_goodfellow2014} is depicted in Fig.~\ref{fig:GAN_basic}.

The architecture of GANs is comprised from two individual components: A discriminator ($D$) and a generator ($G$). $D$ is trained to distinguish between real images from the natural distribution and generated images, while $G$ is trained to craft fake images which fool the discriminator. A random distribution, $\textbf{z} \sim p_{\textbf{z}}$, is given as input to G. The purpose of GANs is to learn the generated samples' distribution, $G(\textbf{z}) \sim p_g$ that estimates the real world distribution $p_r$. GANs are optimized by solving the following min-max optimization problem:
\begin{equation}
\label{gan_minmax}
\min_{G} \max_{D} \mathbb{E}_{\textbf{x} \sim p_r} \log[D(\textbf{x})] + \mathbb{E}_{\textbf{z} \sim p_{\textbf{z}}} \log[1-D(G(\textbf{z}))].
\end{equation}

On the one hand, $D$ aims at predicting $D(\textbf{x})=1$ for real data samples and $D(G(\textbf{z}))=0$ for fake samples. On the other hand, the GAN learns how to fool $D$ by finding $G$ which is optimized on hampering the second term in Eq.~\eqref{gan_minmax}.

On the first iteration, only the discriminator weights $\theta_D$ are updated. We sample a minibatch of $m$ noise samples \{$\textbf{z}^{(1)},...,\textbf{z}^{(m)}$\} from $p_{\textbf{z}}$ and a minibatch of $m$ real data examples \{$\textbf{x}^{(1)},...,\textbf{x}^{(m)}$\} from $p_r$. We then calculate the discriminator's gradients
$$\nabla_{\theta_D} \frac{1}{m} \sum_{i=1}^{m} \left[\log D\left(\textbf{x}^{(i)}\right) + \log\left(1-D\left(G\left(\textbf{z}^{(i)}\right)\right)\right) \right],$$
and update the discriminator weights, $\theta_D$, by ascending this term.
On the second iteration, only the generator's weights $\theta_G$ are updated. We sample a minibatch of $m$ noise samples \{$\textbf{z}^{(1)},...,\textbf{z}^{(m)}$\} from $p_{\textbf{z}}$, and calculate the generator's gradients
\begin{equation}
\label{G_optim}
\nabla_{\theta_G} \frac{1}{m} \sum_{i=1}^{m} \log\left(1-D\left(G\left(\textbf{z}^{(i)}\right)\right)\right),
\end{equation}
and update the generator weights, $\theta_G$, by descending  this term. \citet{GAN_goodfellow2014} showed that under certain conditions on $D$, $G$, and the training procedure, the distribution $p_{\textbf{z}}$ converges to $p_r$.

\section{GAN Advantages and Problems}
\label{GAN Advantages and Problems}
Since their first introduction in 2014, GANs have attracted a growing interest all over the academia and industry, thanks to many advantages over other generative models (mainly Variational Auto-encoders (VAEs) \cite{VAE}):

\begin{itemize}
\item \textbf{Sharp images}: GANs produce sharper images than other generative models. The images at the output of the Generator look more natural and with better quality than images generated using VAEs, which tend to be blurrier.
\item \textbf{Configurable size}: The latent random variable size is not restricted, enriching the generator search space. 
\item \textbf{Versatile generator}: The GAN framework can support many different generator networks, unlike other generative models that may have architectural constraints. VAEs, for example, enforce using a Gaussian at the generator's first layer.
\end{itemize}

The above advantages make GANs very attractive in the deep learning community, achieving state-of-the-art results in a variety of domains, and generating very natural images. Yet, the original GAN suffers from three major problems that are described in detail below. 
We first present a short summary of them:

\begin{itemize}
\item \textbf{Mode collapse}: During the synchronized training of the generator and discriminator, the generator tends to learn to produce a specific pattern (mode) which fools the discriminator. Although this pattern minimizes Eq.~\eqref{gan_minmax}, the generator does not cover the full distribution of the dataset. 
\item \textbf{Vanishing gradients}: Very frequently the discriminator is trained "too well" to distinguish real images from adversarial images; in this scenario, the training step of the generator back propagates very low gradients, which does not help the generator to learn.
\item \textbf{Instability}: The model ($\theta_D$ or $\theta_G$) parameters fluctuate, and generally not stable during the training. The generator seldom achieves a point where it outputs very high-quality images.
\end{itemize}

\subsection{Mode collapse}
\begin{figure}[h]
\centering
\includegraphics[width=\linewidth]{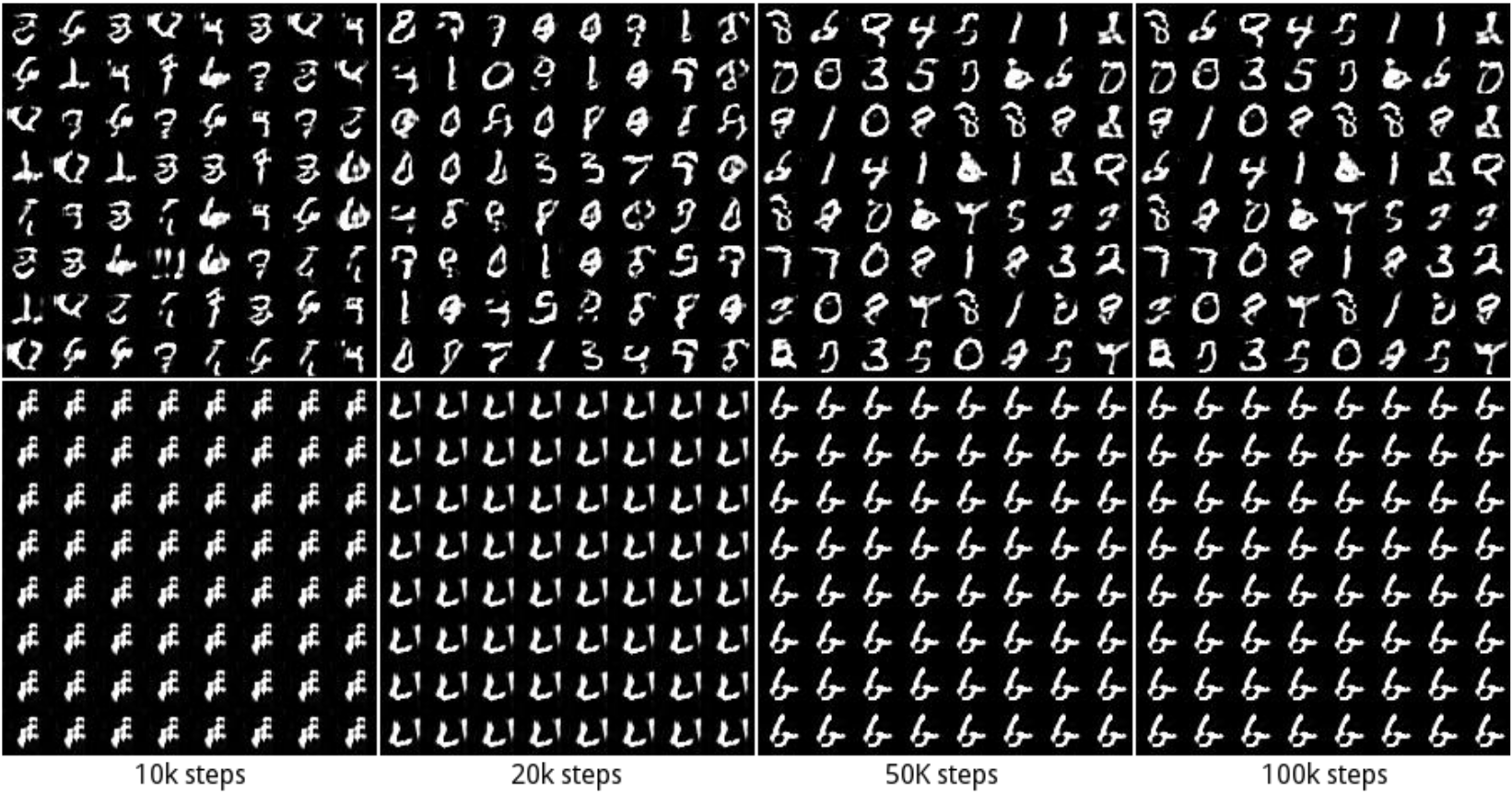}
\caption{Example of the mode collapse problem in GANs. The top row shows a training without mode collapse, where all MNIST modes (digits) are generated. The bottom row shows a bad training with mode collapse, where the generator outputs only the digit '6'. Figure was taken from \cite{DBLP:conf/iclr/MetzPPS17}.}
\label{fig:mode_collapse}
\end{figure}

Data distributions are multi-modal, meaning that every sample is classified (usually) to only one label. For example, in MNIST there are ten classes of digits (or modes) labeled from '0' to '9'.
Mode collapse is the phenomenon where the generator only yields a small subset of the possible modes. In Fig.~\ref{fig:mode_collapse} you can observe generated MNIST images by two different GANs. The top row shows the training of a "good" GAN, not suffering from mode collapse. It generates every kind of mode (digit type) throughout the training. The bottom row exhibits a GAN training with mode collapse, generating just the digit '6'.

\subsection{Vanishing gradients}
\label{Vanishing gradients}
GANs often suffer from training instability, where $D$ performs very well and $G$ does not get a chance to train a good distribution. We turn to provide some mathematical observations and understanding of why training a generator $G$ is extremely hard when the discriminator $D$ is close to optimal.

The global optimality, stated in \cite{GAN_goodfellow2014}, is defined when $D$ is optimized for any given $G$. The optimal $D$ is achieved when its derivative for Eq.~\eqref{gan_minmax} equals 0:
\begin{equation}
\begin{gathered}
\label{D_derivative}
\frac{p_r(\textbf{x})}{D(\textbf{x})} - \frac{p_g(\textbf{x})}{1-D(\textbf{x})} = 0, \\
D^*(\textbf{x}) = \frac{p_r(\textbf{x})}{p_r(\textbf{x}) + p_g(\textbf{x})},
\end{gathered}
\end{equation}
where \textbf{x} is the real input data, $D^*(\textbf{x})$ is the optimal discriminator, and  $p_r(\textbf{x})$/$p_g(\textbf{x})$ is the distribution of the real/generated data, respectively, over the real data \textbf{x}.
If we substitute the optimal discriminator $D^*(\textbf{x})$ into Eq.~\eqref{gan_minmax}, we can visualize the loss for the generator $G$:
\begin{equation}
\label{GAN_loss_optimized}
\begin{aligned}
\mathbb{L}_G = \mathbb{E}_{\textbf{x} \sim p_r} \log\frac{p_r(\textbf{x})}{\frac{1}{2}[p_r(\textbf{x}) + p_g(\textbf{x})]} + \mathbb{E}_{\textbf{x} \sim p_g} \log\frac{p_g(\textbf{x})}{\frac{1}{2}[p_r(\textbf{x}) + p_g(\textbf{x})]} - 2 \cdot \log2.
\end{aligned}
\end{equation}

Before we continue any further, we mention two important metrics for probability measurement. The first one is the Kullback-Leiblar (KL) divergence:
\begin{equation}
\label{KL}
KL(p_1||p_2) = \mathbb{E}_{\textbf{x} \sim p_1}\log\frac{p_1}{p_2},
\end{equation}
which measures how much the distribution $p_2$ differs from the distribution $p_1$. Note that this metric is not symmetrical, i.e., $KL(p_1||p_2) \neq KL(p_2||p_1)$. A symmetrical metric is Jensen-Shannon (JS) divergence, defined as:
\begin{equation}
\label{JS}
JS(p_1||p_2) = \frac{1}{2}KL(p_1||\frac{p_1+p_2}{2}) + \frac{1}{2}KL(p_2||\frac{p_1+p_2}{2}).
\end{equation}

\begin{figure}
\centering
\includegraphics[width=\linewidth]{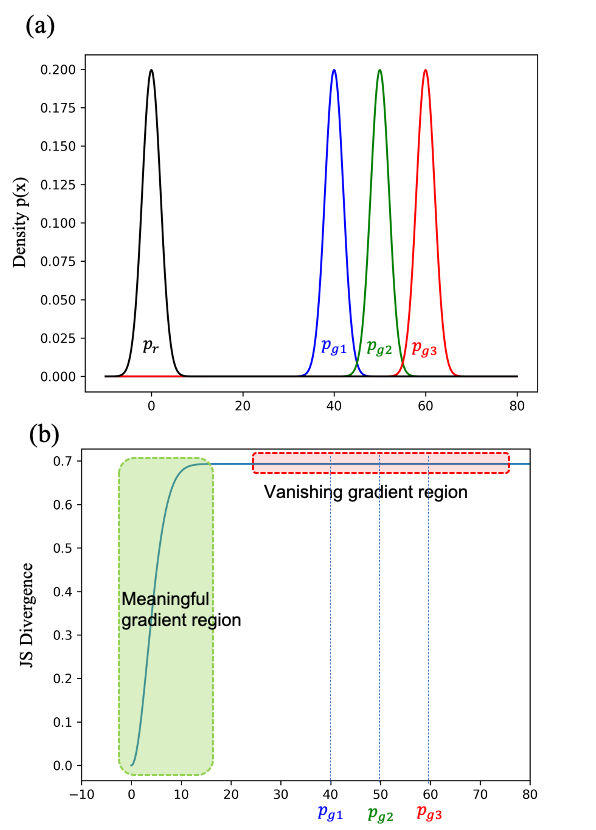}
\caption{Vanishing gradients in GANs. (a) An example of a real image distribution $p_r(\textbf{x})$, a Gaussian with zero mean, with three more different Gaussian distributions: $p_{g1}$, $p_{g2}$, and $p_{g3}$. (b) Calculating the JS divergence measure between $p_r(\textbf{x})$ and a Gaussian distribution with mean from 0 to 80. When training with optimal discriminator ($D^*(\textbf{x})$), the generator $G$ minimizes the loss in Eq.~\eqref{JS}, pushing $p_g(\textbf{x})$ left towards $p_r(\textbf{x})$, alas, it could take a very long time due to diminished gradients when being far from $p_r(\textbf{x})$.}
\label{fig:JS_div}
\end{figure}

Back to our GAN with the optimal $D$, Eq.~\eqref{GAN_loss_optimized} shows that the GAN loss function can be reformulated as:
\begin{equation}
\label{GAN_loss_optimized_reform}
\mathbb{L}_G = 2 \cdot JS(p_r||p_g) -2 \cdot \log2,
\end{equation}
which shows that for $D^*$, the generator loss turns to be a minimization of the JS divergence between $p_r$ and $p_g$.
The relation between GAN training and the JS divergence may explain its instability. To understand this, view Fig.~\ref{fig:JS_div}, which shows an example for JS divergences of different distributions. In Fig.~\ref{fig:JS_div}(a) we see the real image distribution $p_r$, as a Gaussian with zero mean, and we consider three examples for generated image distributions: $p_{g1}$, $p_{g2}$, and $p_{g3}$.
Fig.~\ref{fig:JS_div}(b) plots the $JS(p_r(\textbf{x}), p_q(\textbf{x}))$ measure between $p_r$ and some $p_q$ distribution, where the mean of $p_q$ ranges from $0$ to $80$. As shown in the red box, the gradient of the JS divergence vanishes after a mean of 30. In other words, when the discriminator is close to optimal ($D^*(\textbf{x})$), and we try to train a "poor" generator with $p_g$ far from the real distribution $p_r$, training will not be feasible due to extremely low gradients. This is prominent especially in the beginning of the training where $G$ weights are randomized.

Due to the vanishing gradient problem, Goodfellow \textit{et al.} \cite{GAN_goodfellow2014} proposed a change to the original adversarial loss. Instead of minimizing $\log(1-D(G(\textbf{z})))$ (as in Eq.~\eqref{G_optim}), they suggest to maximize $\log(D(G(\textbf{z})))$. The latter cost function yields the same fixed point of $D$ and $G$ dynamics but maintains higher gradients early in the training when the distributions $p_r$ and $p_g$ are far from each other. On the other hand, this training strategy promotes mode collapse, as we show next.

With an optimal discriminator $D^*$, $KL(p_g||p_r)$ can be reformulated as:
\begin{equation}
\label{KL_pg_pr}
\begin{split}
KL(p_g||p_r) &= \mathbb{E}_{\textbf{x} \sim p_g}\log\frac{p_g(x)/(p_r(x)+p_g(x))}{p_r(x)/(p_r(x)+p_g(x)))}, \\
             &= \mathbb{E}_{\textbf{x} \sim p_g}\log\frac{1-D^*(x)}{D^*(x)}, \\
             &= \mathbb{E}_{\textbf{x} \sim p_g}\log[1-D^*(x)] - \mathbb{E}_{\textbf{x} \sim p_g}\log[D^*(x)].
\end{split}
\end{equation}
If we switch the order of the two sides in Eq.~\eqref{KL_pg_pr}, we get:
\begin{equation}
\label{KL_reverse}
\begin{split}
- \mathbb{E}_{\textbf{x} \sim p_g}\log[D^*(x)] &= KL(p_g||p_r) - \mathbb{E}_{\textbf{x} \sim  p_g}\log[1-D^*(x)], \\
 &= KL(p_g||p_r) - 2 \cdot JS(p_r||p_g) + 2 \cdot \log2 + \mathbb{E}_{\textbf{x} \sim  p_r}\log[D^*(x)].
\end{split}
\end{equation}

\begin{figure}[H]
\centering
\includegraphics[width=\linewidth]{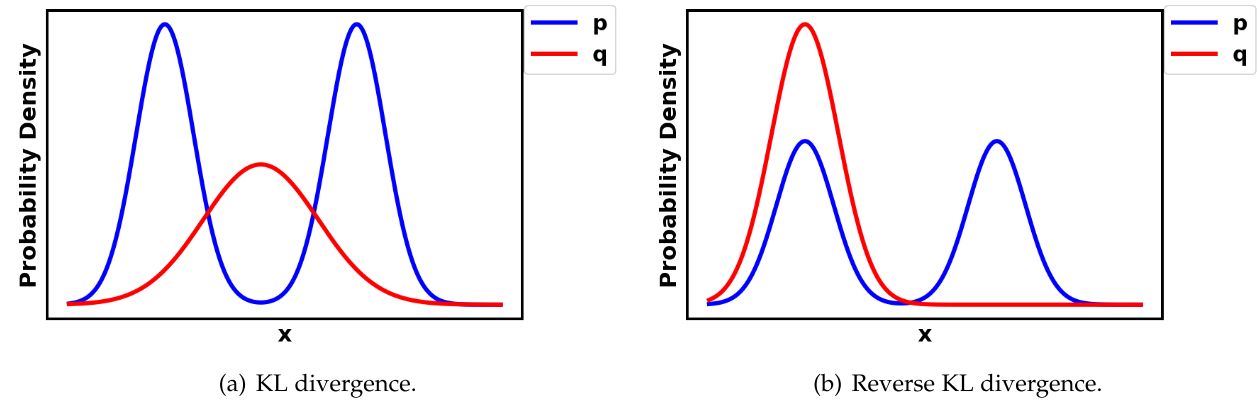}
\caption{Optimized distribution $q$ for (a) minimizing KL divergence $KL(p||q$) and for (b) minimizing reverse KL divergence $KL(q||p)$. Figure is taken from \cite{DBLP:journals/corr/abs-1906-01529}.}
\label{fig:reversed_KL}
\end{figure}

The alternative loss for $G$ is thus only affected by the first two terms (the last two terms are constant). Since $JS(p_r||p_g)$ is bounded by $[0, \log2]$ (see Fig.~\ref{fig:JS_div}(b)), the loss function is dominated by $KL(p_g||p_r)$, which is also called the reverse KL divergence. since $KL(p_g||p_r)$ usually does not equal $KL(p_r||p_g)$, the optimized $p_g$ by the reversed KL is totally different than $p_g$ optimized by the KL divergence. Fig.~\ref{fig:reversed_KL} shows the difference of the two optimization where the distribution $p$ is a mixture of two Gaussians, and $q$ is a single Gaussian. When we optimize for $KL(p_r||p_g)$, $q$ averages all of $p$ modes to hit the mass center (Fig.~\ref{fig:reversed_KL}(a)). However, for the reverse KL divergence optimization, $q$ distribution chooses a single mode (Fig.~\ref{fig:reversed_KL}(b)), which will cause a mode collapse during training.

In summary, using the original $G$ loss from Eq.~\eqref{gan_minmax} will result in vanishing gradients for $G$, and using the other loss in Eq.~\eqref{KL_reverse} will result in a mode collapse. These problems are inherent within the GAN loss and thus cannot be solved using sophisticated architectures. In Section~\ref{losses} we discuss other loss functions for GANs, which solve these problems.

\subsection{Instability and Image quality}
\begin{figure}[h!]
\centering
\includegraphics[width=\linewidth]{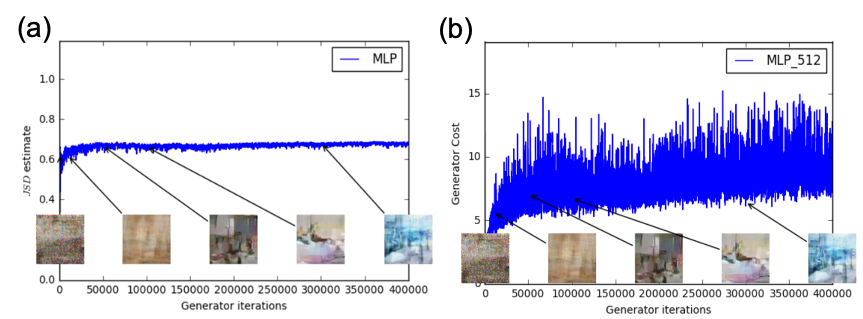}
\caption{Training instability in GANs. (a) $JS$ distance metric in GAN training using the loss in Eq.~\eqref{gan_minmax}. This quantity correlates poorly to the generated image quality, saturating to $\log2 \approx 0.69$, the highest value taken by the $JS$ distance. (b) Generator loss during training using a different generator cost (maximizing $\log(D(G(\textbf{z})))$ instead of minimizing $\log(1-D(G(\textbf{z})))$) showing increasing error without a significant improvement in image quality. Plots are taken from \cite{WGAN}}
\label{fig:G_loss_training}
\end{figure}

Early GAN models which use the G losses described above \Big($\log\big(1-D(G(\textbf{z}))\big)$ and $-\log\big(D(G(\textbf{z}))\big)$\Big) exhibit great instability in their cost values during training. Arjovsky \textit{et al.} \cite{WGAN} experimented with these two $G$ losses and claimed that in both cases the gradients cause instability to the GAN training. The loss in Eq.~\eqref{gan_minmax} stays constant after the first training steps (Fig.~\ref{fig:G_loss_training}(a)), and the loss in Eq.~\eqref{KL_pg_pr} fluctuates during the entire training (Fig.~\ref{fig:G_loss_training}(b)). In both cases, they did not find a significant correlation between the calculated loss and the generated image quality. In other words, it is very difficult to predict when during the training the generator actually produces good quality images, and the only way to get a good generator is to stop the training and manually visualize many generated images.

Using the above two G losses in Eq.~\eqref{gan_minmax} and Eq.~\eqref{KL_pg_pr} yield poor quality images compared to modern GAN models. In the following sections we will cover more sophisticated losses that enhance the generator's resolution and image size.

\subsection{Problems: summary}
\begin{figure}[h!]
\centering
\includegraphics[width=\linewidth]{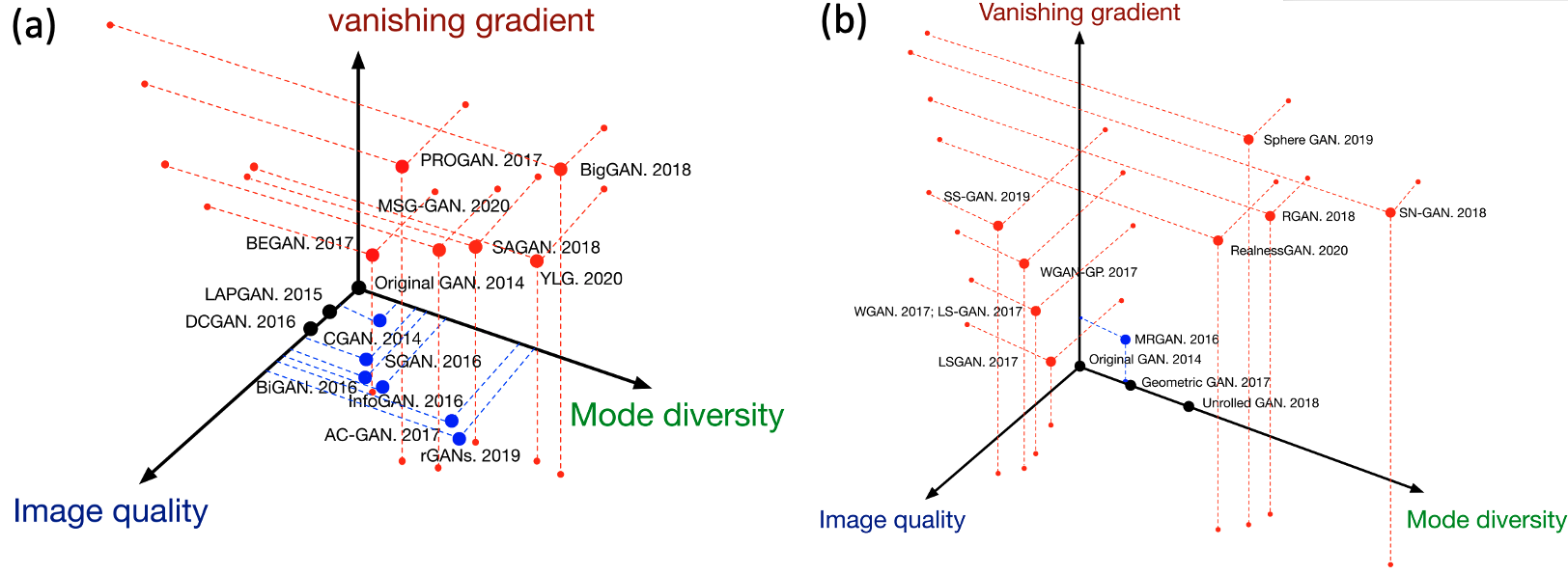}
\caption{Recent GAN models which solve the original GAN's problems: Mode diversity (collapse), vanishing gradients, and image quality. (a) A subset of architecture-variant GANs, (b): A subset of loss-variant GANs. Larger axis values indicate better performance. Red points indicate the model improves all three challenges, blue points improve two, and black points improve only one. Figures are taken from \cite{GANS_review_Wang}.}
\label{fig:problems_summary}
\end{figure}

The original GAN model and loss proposed by Goodfellow \textit{et al.} \cite{GAN_goodfellow2014} suffer from three inherent challenges: (1) Mode collapse; (2) Vanishing gradients; and (3) Image quality. Follow-up works improve the performance of GANs on one or more of these problems by using different architectures for $D$ or $G$, modifying the cost function, and more. A subset of architecture-variant and loss-variant GANs are portrayed in Fig.~\ref{fig:problems_summary}(a),(b), respectively. A sample of some recent prominent GAN models are presented in the following sections.

\section{Improved GAN architectures}
Many types of new GAN architectures have been proposed since 2014 (\cite{BEGAN,BigGAN,LAPGAN,karras2018progressive,DCGAN,SAGAN,StyleGAN,feigin2020gael}). Different GAN architectures were proposed for different tasks, such as image super-resolution \cite{SRGAN} and image-to-image transfer \cite{CycleGAN,park2019SPADE}. In this section we present some of these models, which improved the performance on image quality, vanishing gradients, and mode collapse, compared to the original GAN.

\subsection{Semi-supervised GAN (SGAN)}
\label{SGAN}
Semi-supervised learning is a promising research field between supervised learning and unsupervised learning. In supervised learning, each data is labeled, and in unsupervised learning, no labels are provided; Semi-supervised learning has labels only for a small subset of the training data, and no annotations for the rest of the data, like many real-world problems.

\begin{figure}[H]
\centering
\includegraphics[width=\linewidth]{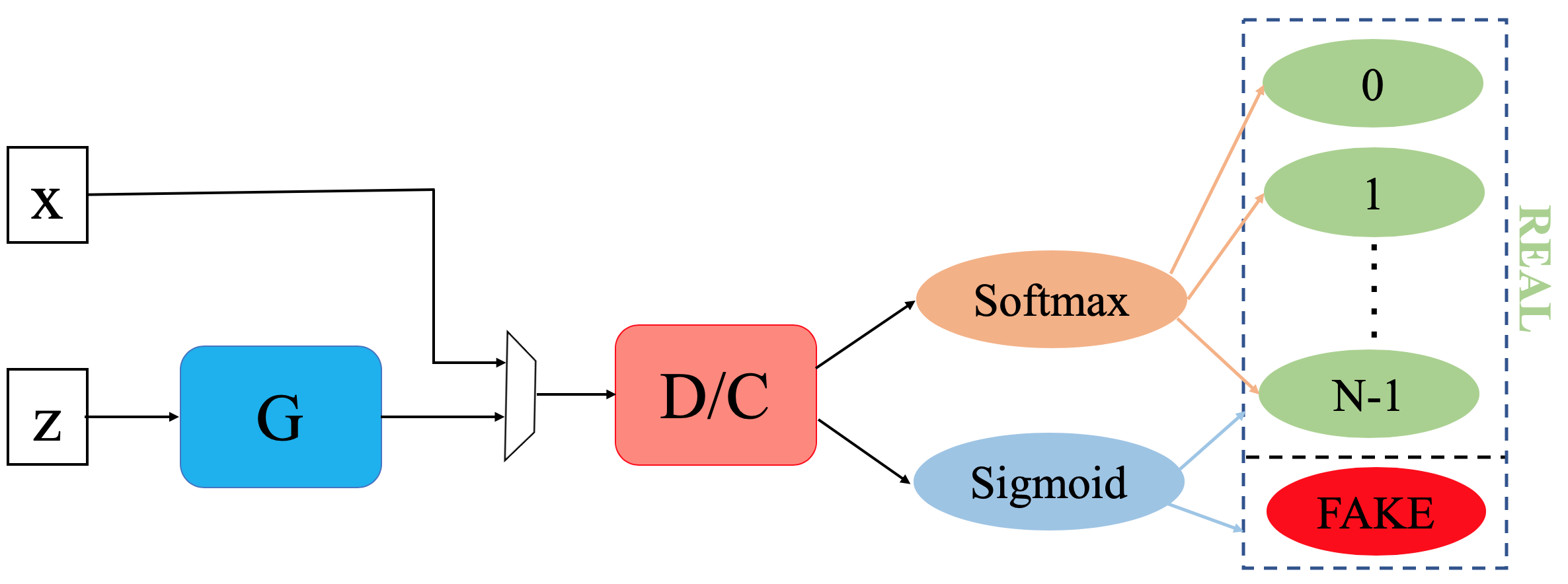}
\caption{SGAN architecture.}
\label{fig:SGAN}
\end{figure}

SGAN \cite{SGAN} extended the original GAN learning to the semi-supervised context
by adding to the discriminator network an additional task of classifying the image labels (Fig.~\ref{fig:SGAN}). The generator's architecture is the same as before. In SGAN the discriminator utilizes two heads, a softmax and a sigmoid. The sigmoid is used to distinguish between real and fake images, and the softmax predicts the images' labels (only for images predicted as real). The results on MNIST showed that both $D$ and $G$ in SGAN are improved compared to the original GAN.

\subsection{Conditional GAN (CGAN)}
\label{CGAN}
\begin{figure}[h!]
\centering
\includegraphics[width=\linewidth]{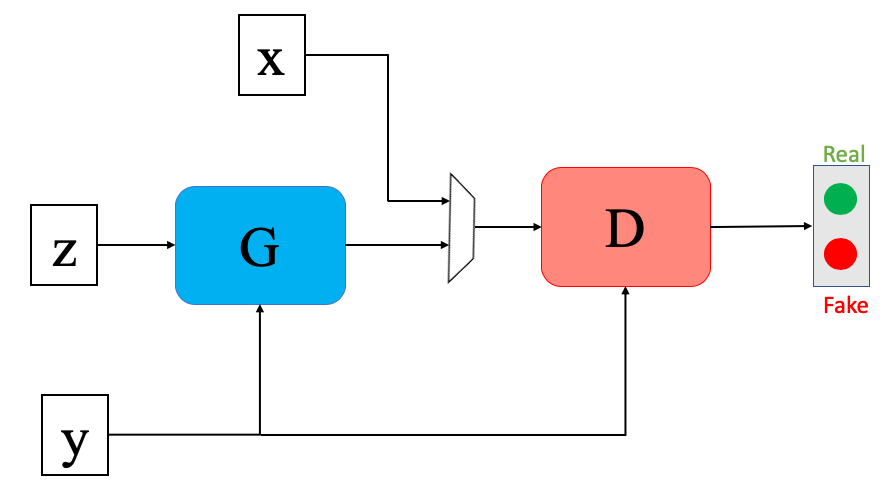}
\caption{CGAN architecture. Both the generator and discriminator are fed with a class label $y$ to condition both the images $x$ and latent variable $z$ (see Eq.~\eqref{CGAN_loss}).}
\label{fig:CGAN}
\end{figure}

CGAN was originally proposed as an extension of the original GAN, where both the discriminator and generator were fed by an additional class of the image \cite{CGAN,pmlr-v70-odena17a}. An illustration of CGAN architecture is shown in Fig.~\ref{fig:CGAN}. In this setup the loss function in Eq.~\eqref{gan_minmax} is slightly modified to condition both the real images $x$ and the latent variable $z$ on $y$ (the label):
\begin{equation}
\label{CGAN_loss}
\min_{G} \max_{D} \mathbb{E}_{\textbf{x} \sim p_r} \log[D(\textbf{x}|\textbf{y})] + \mathbb{E}_{\textbf{z} \sim p_{\textbf{z}}} \log[1-D(G(\textbf{z}|\textbf{y}))].
\end{equation}

All values ($x$, $y$, $z$) are first encoded by some neural layers prior to their fusion in the discriminator and generator. This improves the ability of the discriminator to classify real/fake images and enhances the generator's ability to control the modalities of the generated images.
\cite{CGAN} showed that their CGAN architecture used with a language model can handle also multimodal datasets, such as Flicker that contains labeled image data with their particular user tags. They demonstrated that their generator is capable of producing an automatic tagging.

\subsection{Deep Convolutional GAN (DCGAN)}

\begin{figure}[h!]
\centering
\includegraphics[width=\linewidth]{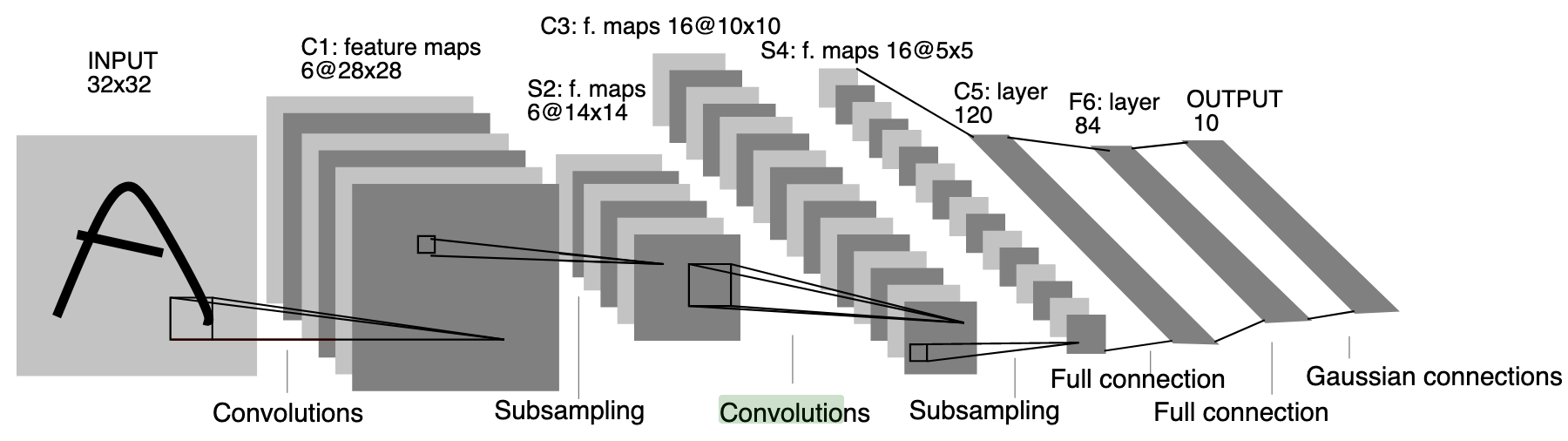}
\caption{LeNet architecture. A popular Convolutional Neural Network (CNN) used for image recognition. Image was taken from \cite{LeCun1999ObjectRW}.}
\label{fig:LeNet}
\end{figure}

Before describing DCGAN, we provide a brief reminder of what is a convolutional neural network. 
\begin{myprop}
Convolutional neural networks (CNNs) were proposed by LeCun \textit{et al.} \cite{LeCun1999ObjectRW}; These  networks consist of trained spatial filters applied on hidden activations throughout their architecture. These networks perform correlations using their trained kernels with a sliding window over images (or hidden activations). This was shown to improve the accuracy on many recognition tasks, especially in computer vision. A very basic and popular CNN network called LeNet is shown in Fig.~\ref{fig:LeNet}.

\end{myprop}

\begin{figure}[h!]
\centering
\includegraphics[width=\linewidth]{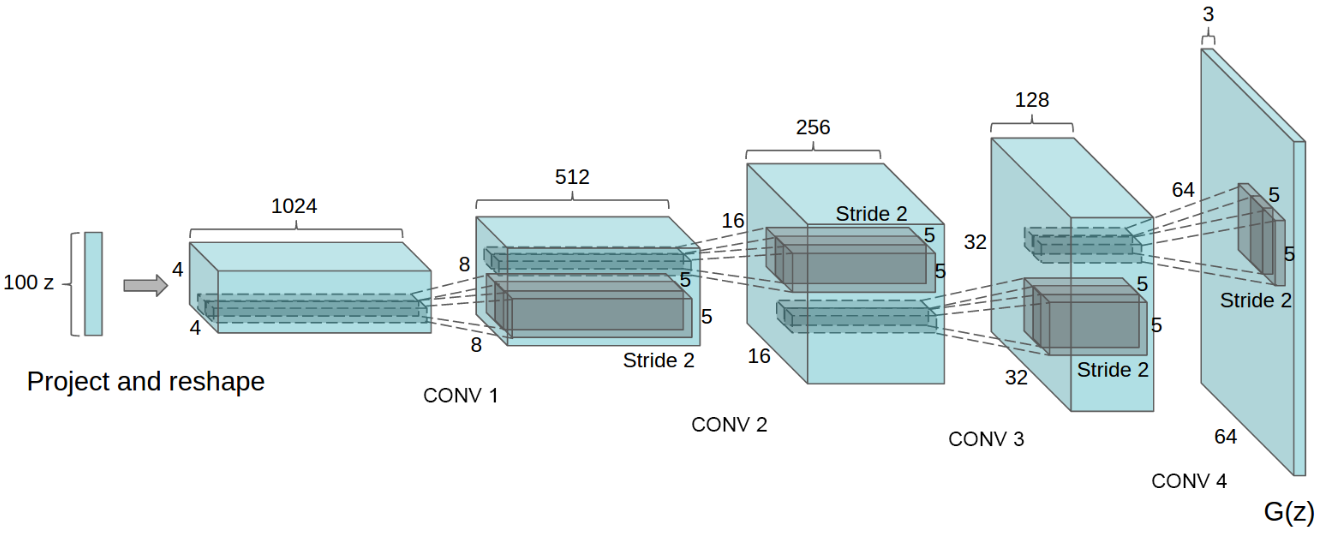}
\caption{DCGAN generator architecture. This generator creates a complex $64\times64$ pixel image from a 100 dimensional uniform distribution $z$. No fully connected or pooling layers are used. Figure is taken from \cite{DCGAN}}
\label{fig:DCGAN}
\end{figure}

DCGAN proposed to create images using solely deconvolutional networks in their generator \cite{DCGAN}. The generator architecture is depicted in Fig.~\ref{fig:DCGAN}. Deconvolutional networks can be conceived as CNNs that use the same components but in reverse, projecting features into the image pixel space. \cite{Zeiler2014VisualizingAU} showed that deconvolutional layers achieve good visualization for CNNs; this allowed the DCGAN generator to create high-resolution images for the first time.

In addition to improved resolution, DCGAN showed better stability during the training thanks to multiple modifications in the original GAN network:
\begin{itemize}
  \item All pooling layers were replaced. In the discriminator, they used convolution kernels with stride > 1, and the generator utilized fractional-strided convolution to increase the spatial size.
  \item Both the discriminator and generator were trained with batch normalization which promotes similar statistics for real images and fake generated images.
  \item The discriminator architecture replaces all normal ReLU activations with Leaky-ReLU \cite{LeakyReLU}; this activation multiplies also the negative kernel output by a small value ($0.2$), to prevent from "dead" gradients to propagate to the generator. The generator architecture used ReLU after every deconvolution layer except the output, which uses Tanh activation.
\end{itemize}

The authors demonstrated empirically that the mere CNN architecture used in DCGAN in not the key contributing factor for the GAN's performance, and the above modifications are crucial.
To show that they measured GANs quality by considering them as feature extractors on supervised datasets, and evaluating the performance of linear models trained on these features (for more information see \cite{DCGAN}). 
DCGAN yields a classification error of 22.48\% on the StreetView House Numbers dataset (SVHN) \cite{SVHN}, whereas a purely supervised CNN with the same architecture achieved a significantly higher 28.87\% test error.

\subsection{Progressive GAN (PROGAN)}
\begin{figure}[h!]
\centering
\includegraphics[width=\linewidth]{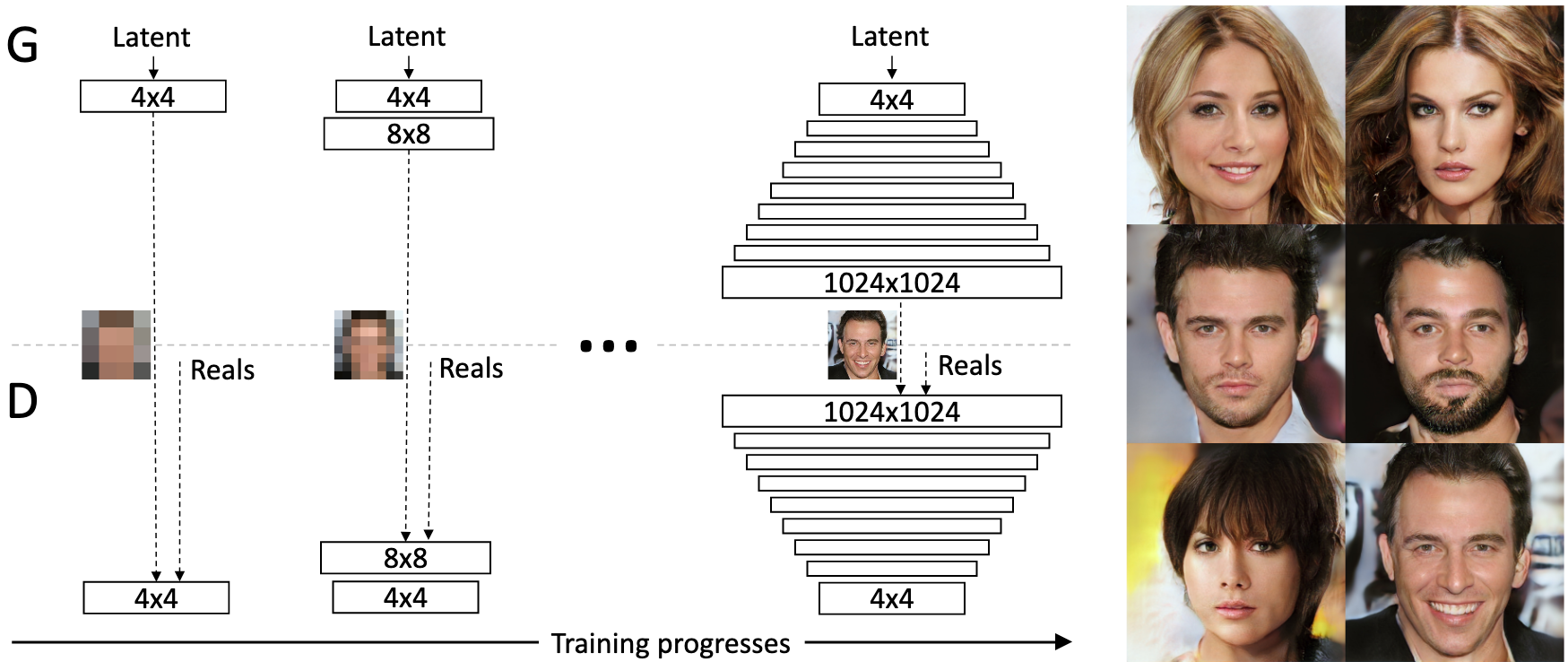}
\caption{Training process for the PROGAN progressive methodology. Training starts with both $G$ and $D$ having a low spatial resolution of $4\times4$ pixels. Every training step adds a new incremented intermediate layer to $D$ and $G$, enhancing the generated images' resolution. All existing layers are trainable throughout the process. The images on the right are examples of generated images using PROGAN at $1024\times1024$. Figure was taken from \cite{karras2018progressive}}
\label{fig:PROGAN}
\end{figure}

PROGAN described a novel training methodology for GANs, involving progressive steps toward the development of the entire network architecture \cite{karras2018progressive}. This progressive architecture uses the idea of progressive neural networks first proposed by Rusu \textit{et al.} \cite{Rusu2016ProgressiveNN}. These architectures are immune to forgetting and can leverage prior knowledge via lateral connections to previously learned features. The progressive training scheme is shown in Fig.~\ref{fig:PROGAN}. First, they trained low resolution $4x4$ pixel images. Next, both $D$ and $G$ grow to include a layer of $8x8$ spatial resolution. This progressive training gradually adds more and more intermediate layers to enhance the resolution, until reaching a high resolution of $1024x1024$ pixel images with the CelebA dataset. All previous layers remain trainable in later steps.

Many state-of-the-art GAN architectures utilize this type of progressive training scheme, and it has resulted in very credible images \cite{karras2018progressive,StyleGAN,BigGAN,Karras2019stylegan2,richardson2020encoding}, and more stable learning for both $D$ and $G$.

\subsection{BigGAN}
Attention in computer vision is a method that focuses the task on an "interesting" region in the image. The self-attention is used to train the network (usually CNN) in an unsupervised manner, teaching it to segment or localize the most relevant pixels (or activations) for the vision task.

BigGAN \cite{BigGAN} has achieved state-of-the-art generation on the ImageNet datasets. Its architecture is based on the Self-attention GAN (SAGAN) \cite{SAGAN}, which employs a self-attention mechanism in both $D$ and $G$, to capture a large receptive field without sacrificing computational efficiency for CNNs \cite{Attention2017}. The original SAGAN architecture can learn global semantics and long-range dependencies for images, thus generating excellent multi-label images based on the ImageNet datasets ($128\times128$ pixels). BigGAN achieved improved performance by scaling up the GAN training: Increasing the number of network parameters ($\times4$) and increasing the batch size ($\times 8$). They achieved better performance on ImageNet with size $128\times128$ and were also able to train BigGAN also on the resolutions $256\times256$ and $512\times512$.

\begin{figure}[h!]
\centering
\includegraphics[width=\linewidth]{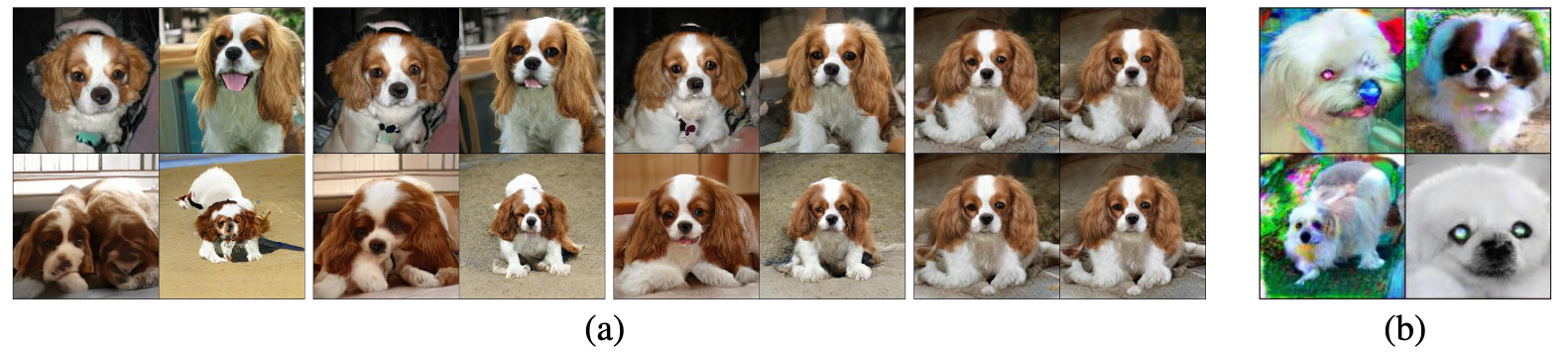}
\caption{Using truncated sampling from the latent space. (a) Trade-off between image quality and image variety. From left to right, the range threshold is set to $2$, $1$, $0.5$, and $0.04$. (b) Saturation artifacts from applying truncated normal distribution to a model training with $z \sim \mathcal{N}(0,\,1)$. Images were taken from \cite{BigGAN}.}
\label{fig:BigGAN_truncation}
\end{figure}

Unlike many previous GAN models, which randomized the latent variable from either $z \sim \mathcal{N}(0,\,1)$ or $z \sim \mathcal{U}(-1,\,1)$, BigGAN uses a simple \textit{truncation trick}. During training $z$ is sampled from $\mathcal{N}(0,\,1)$, but for generating images in inference $z$ is selected from a truncated normal, where values that lie outside the range are re-sampled until falling in the range. This \textit{truncation trick} shows improvement in individual sample quality at the cost of a reduction in overall sample variety, as shown in Fig.~\ref{fig:BigGAN_truncation}(a). Notice though that using this different inference sampling trick as is may not generate good images with some large models, producing some saturation artifacts (Fig.~\ref{fig:BigGAN_truncation}(b)). To solve this issue, the authors proposed to use Orthogonal Regularization to force $G$ to be more amenable to truncation, making it smoother so that the full lateral space will map to good generated images.

\subsection{StyleGAN}
\label{StyleGAN}
\begin{figure}[H]
\centering
\includegraphics[width=\linewidth]{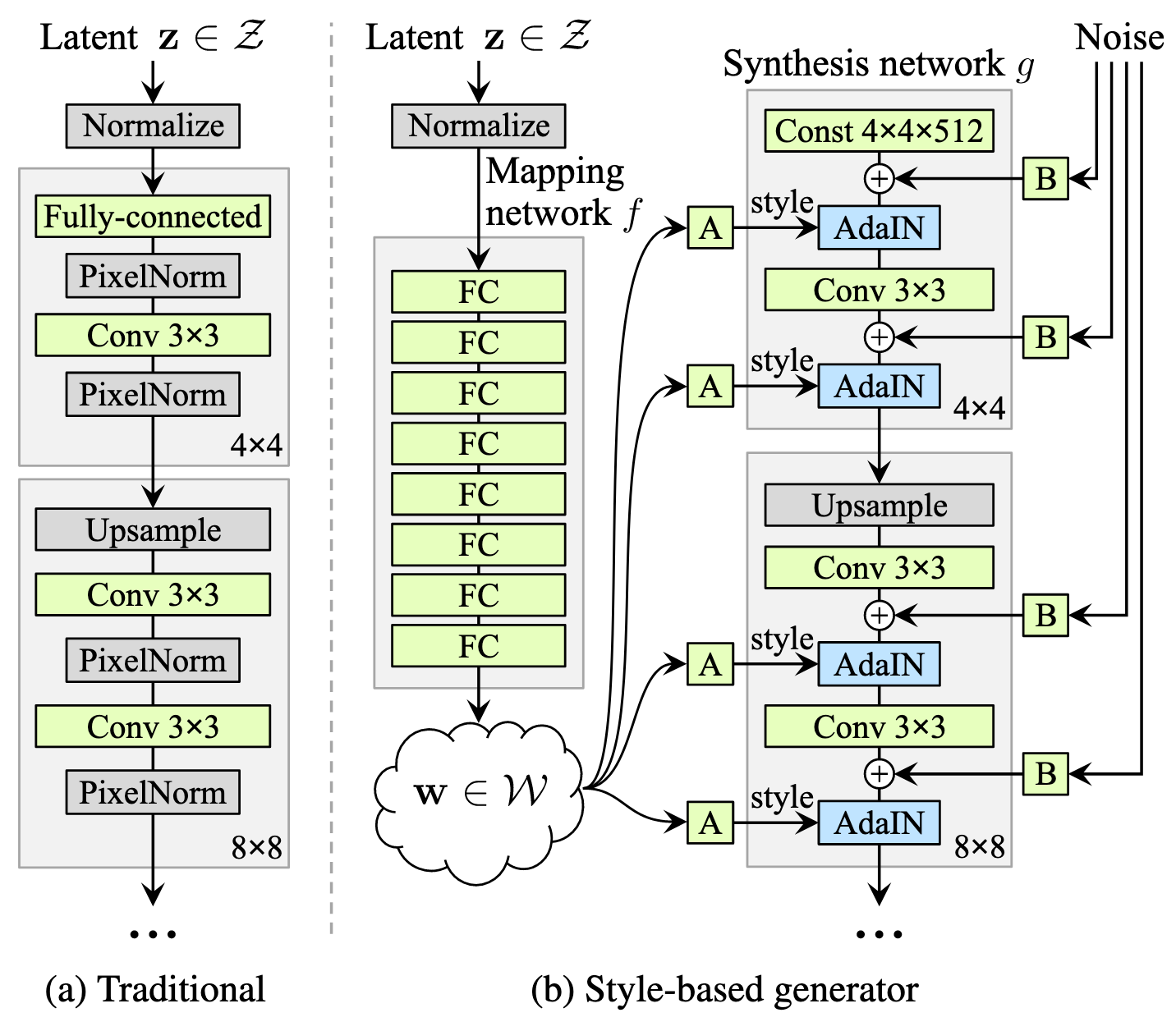}
\caption{Comparison between a traditional generator (left) to the StyleGAN generator (right). Unlike the standard generator which selects a random latent variable only in its input, in StyleGAN generator the latent input is mapped to an intermediate latent space $\mathcal{W}$, which then controls the generator through adaptive instance normalization (AdaIN). Also, Gaussian noise is injected after every convolution layer. "A" corresponds to a learned affine transform, and "B" applies learned scaling factors to the input noise. Figure is taken from \cite{StyleGAN}.}
\label{fig:StyleGAN}
\end{figure}

StyleGAN \cite{StyleGAN} proposed an alternative generator architecture for GANs. Unlike the traditional $G$ architecture that samples a random latent variable at its input, their architecture starts from a learned constant input and adjust the characteristics of the images in every convolutional layer along $G$ using different, learned latent variables. Additionally, they inject learned noise directly throughout $G$ (see Fig.~\ref{fig:StyleGAN}).

\begin{figure}[H]
\centering
\includegraphics[width=\linewidth]{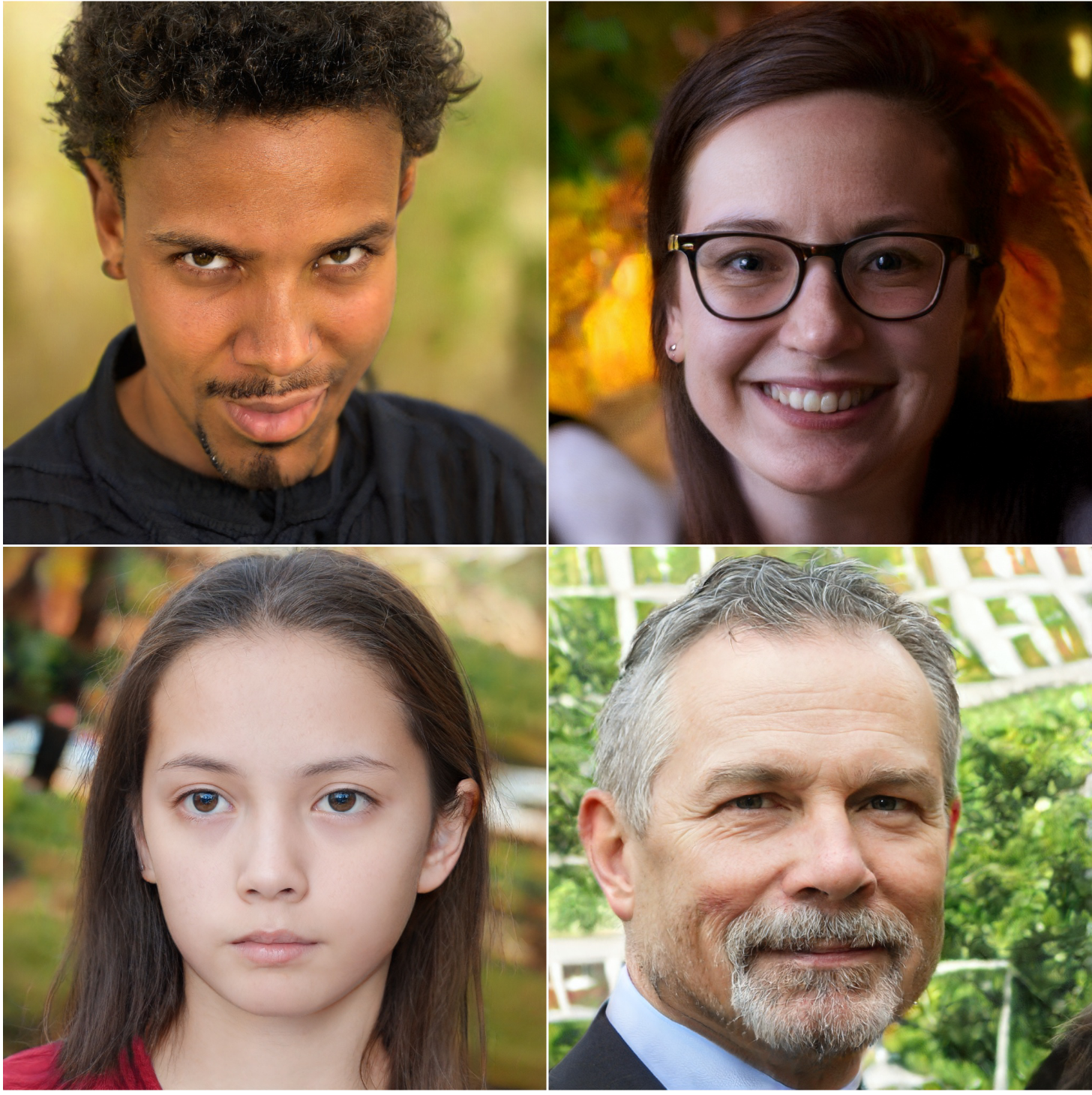}
\caption{Selected face images generated using the StyleGAN2 architecture (an improved version of the original StyleGAN), trained on the FFHQ dataset. Figure from \cite{Karras2019stylegan2}.}
\label{fig:StyleGAN_examples}
\end{figure}

This architectural change leads to automatic, unsupervised separation of high-level attributes (e.g., pose and identity) from low-level variations (e.g., freckles, hair) in the generated images. StyleGAN did not modify the discriminator or the loss function. They used the same $D$ architecture as in \cite{karras2018progressive}. In addition to state-of-the-art quality for face image generation, StyleGAN demonstrates a higher degree of latent space disentanglement, presenting more linear representations of different factors of variation, turning the GAN synthesis to be much more controllable. Recently a more advanced styleGAN architecture has been proposed \cite{Karras2019stylegan2,Rinon21SWAGAN,Karras2021stylegan3}. Some generated examples are presented in Fig.~\ref{fig:StyleGAN_examples}. One may also use StyleGAN for image editing by calculating the latent vector of a given input image in the styleGAN (this operation is known as styleGAN inversion) and then manipulating this vector for editing the image \cite{xia2021gan,richardson2020encoding,Abdal19Image2StyleGAN,Tov21Designing,Abdal_2020_CVPR,shen2020interfacegan,shen2021closedform,patashnik2021styleclip}.

\section{Improved GAN objectives}
\label{losses}
As described in Section~\ref{Vanishing gradients}, the original GAN's min-max loss (Eq.~\eqref{gan_minmax}) promotes mode collapse and vanishing gradient phenomena. This section presents selected loss functions and regularizations that remedy these problems and also improves the image quality. This is just a partial list and other loss functions and regularization methods exist such as the least square GANs \cite{Mao_2017_ICCV} or optimal transport models \cite{Peyre19Computational}.

\subsection{Wasserstein GAN (WGAN)}  
WGAN \cite{WGAN} has solved the vanishing gradient and mode collapse problems of the original GAN by replacing the cost in Eq.~\eqref{gan_minmax} with the Earth Mover (EM) distance, which is known also as the Wasserstein distance and is defined as:
\begin{equation}
\label{WGAN_EM}
W(p_r,p_g)=\inf_{\gamma \in \prod (p_r,p_g)} \mathbb{E}_{(x,y) \sim \gamma} [||x-y||],
\end{equation}
where $\prod (p_r,p_g)$ denotes the set of all joint distributions $\gamma(x,y)$ whose marginals are $p_r$ and $p_g$, respectively. The EM distance is therefore the minimum cost of transporting "mass" in converting distribution $p_r$ into the distribution $p_g$.

\begin{figure}[h!]
\centering
\includegraphics[width=\linewidth]{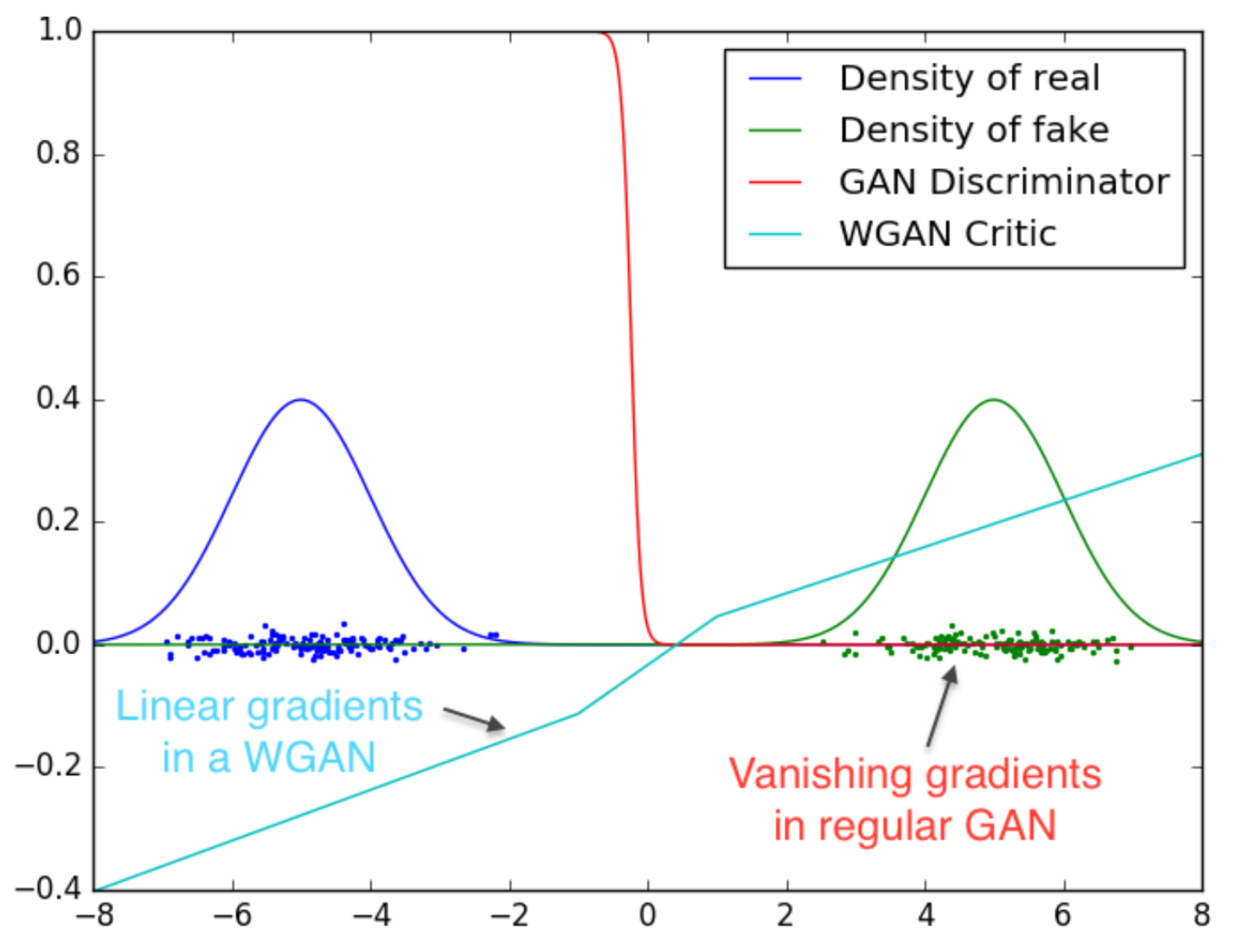}
\caption{Differentiating two Gaussian distributions using optimal discriminator $D^*$. The original GAN cost saturates (red line), resulting in vanishing gradients, whereas WGAN cost objective (blue line) yields measurable gradients. Plot was taken from \cite{WGAN}.}
\label{fig:WGAN_grads}
\end{figure}

Unlike KL and JS distances, EM is capable of indicating distance even when the $p_r$ and $p_g$ distributions are far from each other; EM is also continuous and thus provides useful gradients for training $G$.  However, the infimum in Eq.~\eqref{WGAN_EM} is highly intractable, so the authors estimate the EM cost with:
\begin{equation}
\label{WGAN_EM2}
\max_{w \sim \mathcal{W}} \mathbb{E}_{\textbf{x} \sim p_r}[f_w(\textbf{x})] - \mathbb{E}_{\textbf{z} \sim p_z}[f_w(G(\textbf{z}))],
\end{equation}
where $\{f_w\}_{w\in\mathcal{W}}$ is a parameterized family of all functions that are $K$-Lipschitz for some $K$ ($||f||_L\leq K$). Readers are referred to \cite{WGAN} for more details.

The authors proposed to find the best function $f_w$ that maximize Eq.~\eqref{WGAN_EM2} by back-propagating $\mathbb{E}_{\textbf{z} \sim p_z}\big[\nabla_\theta f\big(g_\theta(z)\big)\big]$, where $g_\theta$ are the generator's weights. $f_w$ can be realized by $D$ but constrained to be K-Lipschitz  and $z$ is the lateral input noise for $G$. $w$ in $f_w$ are the discriminator's parameters, and the objective of $D$ is to maximize Eq.~\eqref{WGAN_EM2}, which approximates the EM distance. When $D$ is optimized, Eq.~\eqref{WGAN_EM2} becomes the EM distance, and $G$ is optimized to minimize it:
\begin{equation}
\label{WGAN_G_loss}
- \min_{G} \mathbb{E}_{\textbf{z} \sim p_z}[f_w(G(\textbf{z}))]
\end{equation}
Fig.~\ref{fig:WGAN_grads} compares the gradient of WGAN to the original GAN from two non overlapping Gaussian distributions. It can be observed that WGAN has a smooth and measurable gradient everywhere (blue line), and learns better even when $G$ is not producing good images.

\subsection{Self Supervised GAN (SSGAN)}
We showed in Section~\ref{CGAN} that CGANs can generate natural images. However, they require labeled images to do so, which is a major drawback. SSGANs \cite{SSGAN} exploit two popular unsupervised learning techniques, adversarial training, and self-supervision, bridging the gap between conditional and unconditional GANs.

\begin{figure}[H]
\centering
\includegraphics[width=\linewidth]{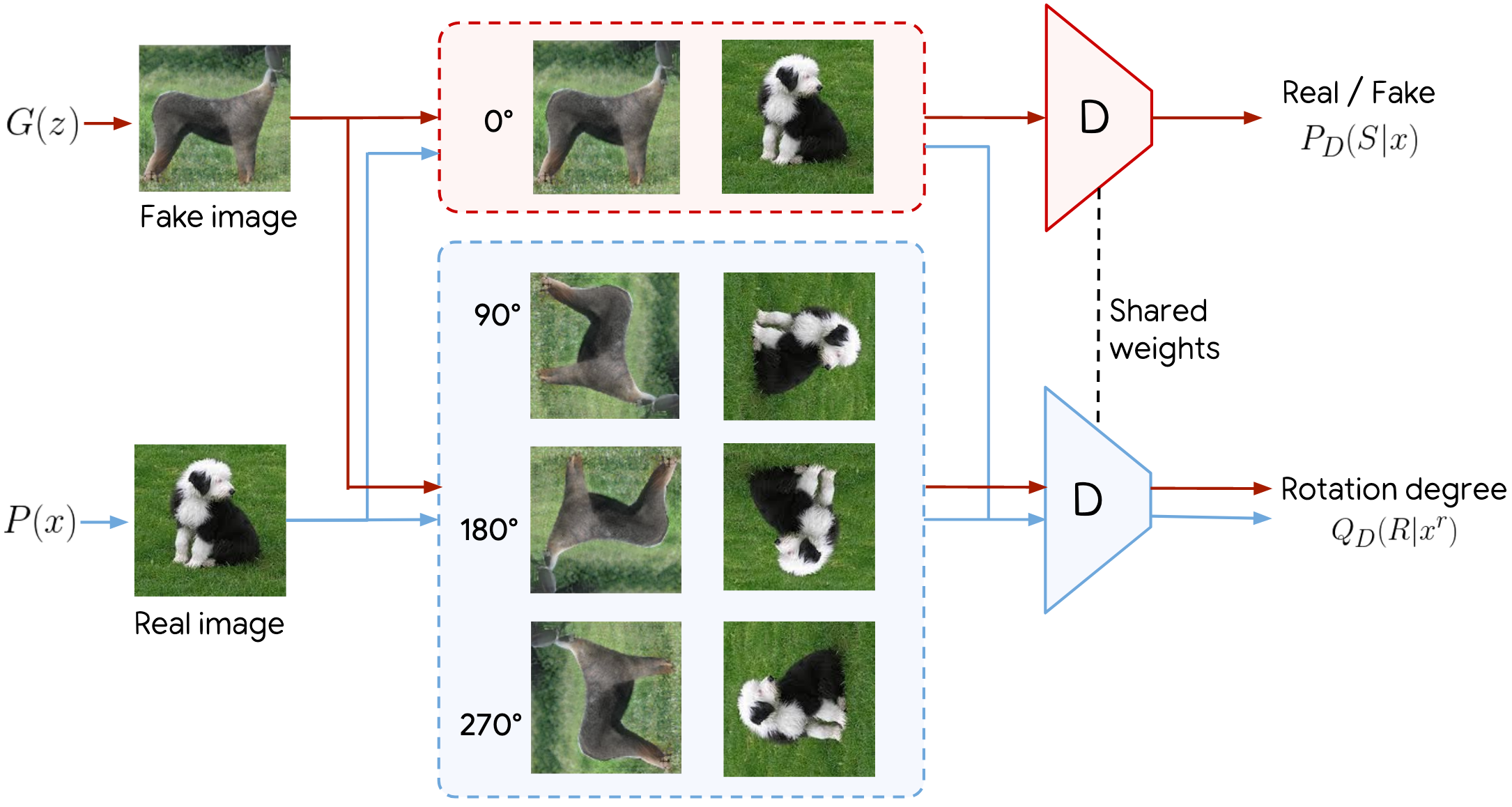}
\caption{Self-Supervised GAN (SSGAN) discriminator. The discriminator, $D$, performs two tasks: Identifying real/fake images (as the original GAN) and rotation classification. Both the real and fake images are rotated by ($[\ang{0}, \ang{90}, \ang{180}, \ang{270}]$), and sent to the rotation degree classifier (in blue). Only the original $\ang{0}$ images are sent to the real/fake classifier (in red). Plot is taken from \cite{SSGAN}.}
\label{fig:SSGAN}
\end{figure}

Neural networks have been shown to forget previous learned tasks \cite{CatastrophicForgetting1999,Kirkpatrick2017OvercomingCF}, and catastrophic forgetting was previously considered as a major cause for GAN training instability. Motivated by the desire to counter the discriminator forgetting, SSGAN adds to the discriminator an additional loss, which enables it to learn useful representations, independently of the quality of the generator. In a self-supervised manner, the authors train a model on a task of prediction a rotation angle ($[\ang{0}, \ang{90}, \ang{180}, \ang{270}]$), as shown in Fig.~\ref{fig:SSGAN}. The objectives of $D$ and $G$ are updated to:
\begin{equation}
\label{SSGAN_loss}
\begin{split}
L_G &= -V(G,D) - \alpha \mathbb{E}_{\textbf{x} \sim P_G}\mathbb{E}_{r \sim \mathbb{R}}[\log Q_D(R=r|\textbf{x}^r)] \\
L_D &= V(G,D) - \beta \mathbb{E}_{\textbf{x} \sim P_{data}}\mathbb{E}_{r \sim \mathbb{R}}[\log Q_D(R=r|\textbf{x}^r)],
\end{split}
\end{equation}
where $V(G,D)$ is the original GAN objective in Eq.~\eqref{gan_minmax}, $P_{data}$ and $P_G$ are the real data and generated data distributions, respectively, $r \in R$ is a rotation selected from a set of all allowed angles ($R = \{\ang{0}, \ang{90}, \ang{180}, \ang{270}\}$). An image \textbf{x} rotated by $r$ degrees is denoted as $\textbf{x}^r$, and $Q(R|\textbf{x}^r)$ is the discriminator's predictive distribution over the angles of rotation of the sample. These new losses enforce $D$ to learn good representation via learning the rotation information in a self-supervised approach.

Using the above scheme, SSGAN achieves good high-quality images, matching the performance of conditional GANs without having access to labeled data.

\subsection{Spectral Normalization GAN (SNGAN)}
SNGAN \cite{SNGAN} proposes to add weight normalization to stabilize the training of the discriminator. Their technique is computationally inexpensive and can be applied easily to existing GANs architectures. In previous works that stabilized GAN training (\cite{NIPS2017_7159,WGAN,Qi2019LossSensitiveGA}) it was emphasized that $D$ should be a K-Lipshitz continuous function, forcing it not to change rapidly. This characteristic of $D$ stabilizes the training of GANs.  SNGAN controls the Lipschitz constant of $D$ by literally constraining the spectral norm of each layer, normalizing each weight matrix $W$ so it satisfies the spectral constraint $\sigma(\textbf{W})=1$ (i.e., the largest singular value of the weight matrix of each layer is 1). This is performed by simply normalizing each layer:
\begin{equation}
\label{SNGAN_spectral_norm}
\bar{\textbf{W}}_{SN}(\textbf{W})=\frac{\textbf{W}}{\sigma({\textbf{W}})},
\end{equation}
where \textbf{W} are the weight parameters of each layer in $D$. This paper proves that this will make the Lipschitz constant of the discriminator function to be bound by $1$, which is important for the WGAN optimization.

SNGAN achieves an extraordinary advance on ImageNet, and better or equal quality on CIFAR-10 and STL-10, compared to the previous training stabilization techniques that include weight clipping \cite{WGAN}, gradient penalty \cite{Wu2019GPGANTR,Mescheder2018ICML}, batch normalization \cite{Ioffe2015BatchNA}, weight normalization \cite{NIPS2016_6114}, layer normalization \cite{Ba2016LayerNormalization}, and orthonormal regularization \cite{Brock2017NeuralPE}.


\subsection{SphereGAN}
\begin{figure}[H]
\centering
\includegraphics[width=\linewidth]{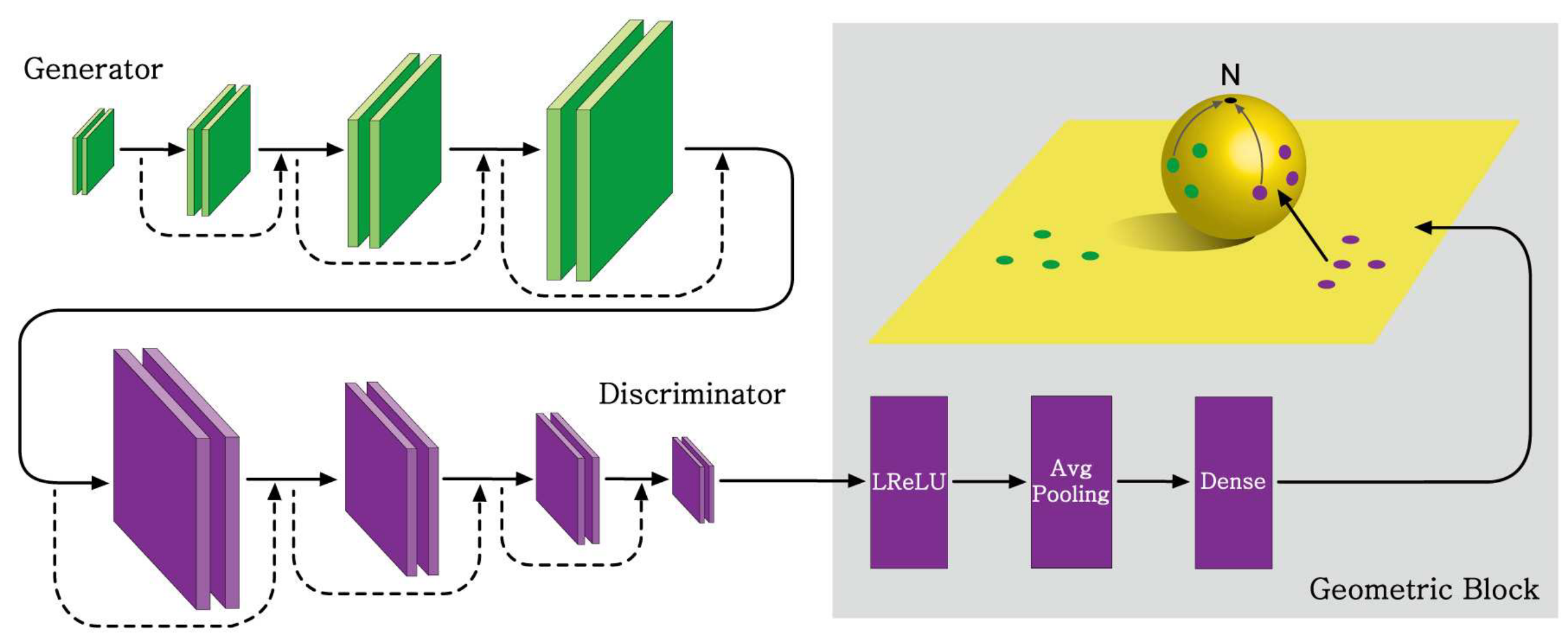}
\caption{Pipeline of Sphere GAN. The generator creates fake data from noise inputs. Real and fake data are fed to the discriminator, which maps the output to an $n$-dimensional Euclidean feature space (yellow plane). Green and purple points on the feature plane correspond to fake and real samples, respectively. The key idea of SphereGAN is remapping the feature points into the $n$-dimensional hypersphere by using geometric transformations. These mapped points are then used to calculate the geometric moments centered at the north pole of the hypersphere ($N$). While the discriminator tries to maximize the moment differences of real and fake samples, the generator tries to interfere with the discriminator by minimizing those moment differences. Figure was taken from \cite{SphereGAN}.}
\label{fig:SphereGAN}
\end{figure}

SphereGAN \cite{SphereGAN} is a novel integral probability metric (IPM)-based GAN, which uses the hypersphere to bound IPMs in the objective function, thus enhancing the stability of the training. By exploiting the information of higher-order statistics of data using geometric moment matching, they achieved more accurate results. The objective function of SphereGAN is defined as
\begin{equation}
\label{SphereGAN_cost}
\min_{G}\max_{D}\sum_rE_x[d^r_s(\textbf{N},D(x))] - \sum_rE_z[d^r_s(\textbf{N},D(G(z)))],
\end{equation}
for $r=1,\dots,R$ where the function $d^r_s$ measures the $r$-th moment distance between each sample and the north pole of the hypersphere, \textbf{N}. Note that the subscript $s$ indicates that $d^r_s$ is defined on $\mathbb{S}^n$. Fig.~\ref{fig:SphereGAN} shows the pipeline of SphereGAN.
By defining IPMs on the hypersphere, SphereGAN can alleviate several constraints that should be imposed on $D$ for stable training, such as the Lipschitz constraints required from conventional discriminators based on the Wasserstein distance.

Unlike conventional approaches that use the Wasserstein distance and add additional constraint terms (see Table 1 in \cite{SphereGAN}), SphereGAN does not need any additional constraints to force $D$ in a desired function space, due to the usage of geometric transformation in $D$.

\section{Data augmentation with GAN}

We turn now to describe how to use GAN for data augmentation.
Data augmentation is a crucial process for tasks lacking large training sets. There are three circumstances which require data augmentation:
\begin{itemize}
\item \textbf{Limited annotations}: Training a large DNN with a very few labeled data in the training set.
\item \textbf{Limited diversity}: When the training data lacks variations. For example, it may not cover diverse illuminations or a variety of appearances. 
\item \textbf{Restricted data}: The database might contain sensitive information, and thus accessing it directly is strictly restricted.
\end{itemize}

The first two scenarios can be solved via supervised learning but they will cost a lot of human effort to enrich the labeled data or to use active learning approaches \cite{10.5555/1622737.1622744}. An alternative approach is to utilize GANs to augment the data. Semi-supervised GAN (SGAN) (see Section~\ref{SGAN}) is a simple example of such a GAN architecture that is able to generate new annotated images and can automatically enrich training data with few labels.

The second case is more common in the real world. Data variability is important for many psychology or neuroscience experiments. For example, a human's EEG is sensitive to different types of face images such as happy, angry, and sad faces \cite{Mavratzakis2016EmotionalFE}. Preparing those stimuli in traditional experiments is time-consuming and costly for researchers. Architectures like StyleGAN \cite{StyleGAN} (see Section~\ref{StyleGAN}) are specialized to generate broad types of face images as a stimulus. More important, these images can be controlled to exhibit specific attributes (e.g. level of happiness or facial textures), and thus enhancing the stimuli variation in the experiment \cite{DBLP:journals/ijon/WangSSWH20}.
Data augmentation can be also helpful in the training of the GANs themselves \cite{Karras2020ada,zhao2020diffaugment}, which allows to train them with only few examples and still get high quality generated data (e.g., images).

The last case is related to unsupervised learning approaches. When a database is restricted to preserve users' privacy, researchers can use GANs to synthesize this sensitive data themselves. For example, Delaney \textit{et al.} employed GANs to generate synthetic ECG signals that resemble real ECG data \cite{Delaney2019SynthesisOR}.


\section{Conclusion}

This chapter has provided just a glimpse of GANs and their various usages. This fast-growing technique has many variants and applications that have not been presented here for the sake of brevity. Yet, we believe that the description of GANs given here provides the reader with the tools to better understand this tool and be able to navigate between the vast amount of recent literature on this topic. It is worth mentioning also normalizing flows \cite{Papamakarios21Normalizing} and score based generative models \cite{song2021scorebased,song2021denoising,pmlr-v139-nichol21a}, which become very popular recently and show competitive performance with GANs.

\clearpage

 \begin{acknowledgement}
We would like to thank Yuval Alaluf, Yotam Nitzan and Ron Mokady for their helpful comments.
 \end{acknowledgement}

%

\bibliographystyle{plainnat}
\bibliography{my_bib.bib}

\end{document}